\documentclass{article}

\usepackage{arxiv}

\usepackage[utf8]{inputenc} 
\usepackage[T1]{fontenc}    
\usepackage{hyperref}       
\usepackage{url}            
\usepackage{booktabs}       
\usepackage{amsfonts}       
\usepackage{nicefrac}       
\usepackage{microtype}      
\usepackage{lipsum}
\usepackage{graphicx}

\usepackage{graphicx}%
\usepackage{multirow}%
\usepackage{amsmath,amssymb,amsfonts}%
\usepackage{amsthm}%
\usepackage{mathrsfs}%
\usepackage[title]{appendix}%
\usepackage{xcolor}%
\usepackage{textcomp}%
\usepackage{manyfoot}%
\usepackage{booktabs}%
\usepackage{algorithm}%
\usepackage{algpseudocode}%
\usepackage{listings}%

\usepackage{amsmath,amssymb,amsfonts}
\usepackage{multirow}
\usepackage{rotating}
\usepackage{booktabs}
\usepackage{subfig}
\usepackage{balance}
\usepackage{acronym}
\usepackage{tabularx}
\usepackage{booktabs}
\usepackage{array}

\newcommand{\refeq}[1]{(\ref{#1})}
\newcommand{\reffig}[1]{Figure~\ref{#1}}

\newcommand{\reftab}[1]{Table~\ref{#1}}

\newcommand{\refsec}[1]{Section~\ref{#1}}

\acrodef{pHRI}{Physical Human-Robot Interaction}
\acrodef{ISO}{International Organization for Standardization}
\acrodef{PFL}{Power and Force Limiting}
\acrodef{SSM}{Speed and Seperation Monitoring}
\acrodef{CTM}{Computed Torque Method}
\acrodef{PD}{Proportional Derivative}
\acrodef{DoF}{Degrees of Freedom}
\acrodef{AR}{Augmented Reality}

\title{Safe Physical Human-Robot Interaction through Variable Impedance Control based on ISO/TS 15066}

\author{
 Armin Ghanbarzadeh \\
  Faculty of Mechanical Engineering\\
  K.N. Toosi University of Technology\\
  Tehran, Iran \\
  \texttt{armingh@email.kntu.ac.ir} \\
   \And
 Esmaeil Najafi \\
  Faculty of Mechanical Engineering\\
  K.N. Toosi University of Technology\\
  Tehran, Iran \\
  \texttt{najafi.e@kntu.ac.ir} \\
}

\begin{document}
\maketitle

\begin{abstract}The successful implementation of \acl{pHRI} in industrial environments depends on ensuring safe collaboration between human operators and robotic devices. This necessitates the adoption of measures that guarantee the safety of human operators in close proximity to robots, without constraining the speed and motion of the robotic systems. This paper proposes a novel variable impedance-based controller for cobots that ensures safe collaboration by adhering to the ISO/TS 15066 safety standard, namely power and force limiting mode, while achieving higher operational speeds. The effectiveness of the proposed controller has been compared with conventional methods and implemented on two different robotic platforms. The results demonstrate the designed controller achieves higher speeds, while maintaining compliance with safety standards. The proposed variable impedance holds significant potential for enabling efficient and safe collaboration between humans and robots in industrial settings.\end{abstract}


\ac{pHRI} refers to the direct physical interaction between humans and robots in shared workspace environments. As robots are becoming increasingly integrated into human-centered workplaces, ensuring safety in \ac{pHRI} is critical. The \ac{ISO} plays a significant role in ensuring safety in \ac{pHRI} by specifying guidelines and requirements for the design, production, and use of robots to guarantee safe human--robot collaboration.

The ISO/TS~15066 standard \cite{ISOTS15066}, for example, provides norms for the safe physical interaction between humans and robots. It covers the design of robots and their components, taking into account their intended use and the environment in which they will be used. The standard also covers the evaluation of the safety of the robot and its components and the documentation required for its safe use. Other \ac{ISO} standards, such as ISO 10218~\cite{ISO102181, ISO102182} and ISO 13482~\cite{ISO13482}, also contribute to the safe design and use of robots in human-centered work environments. These standards, along with others, provide a foundation for the development of safe and effective \ac{pHRI} systems, promoting the integration of robots into human-centered work environments and enhancing human--robot collaboration.

\begin{figure}[t!]
\centering
\subfloat{\includegraphics[width=0.55\linewidth]{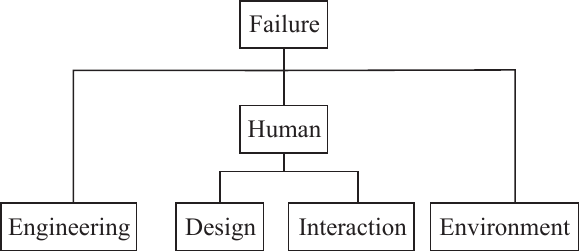}}
\caption{Classification of failures caused by engineering, human operator, or environmental conditions errors, adopted from \cite{ogorodnikova2008methodology}.}
\label{fig:hazards}
\end{figure}

Before attempting to reduce hazards and establish safety, workspace hazards must be accurately identified. In human--robot collaboration, ``workspace hazards'' refers to the potential risks that could result from interactions between people and robots in shared workspaces. A categorization of these dangers is presented in \reffig{fig:hazards}. These hazards can include physical harm to human workers, such as crush injuries or cuts from sharp edges, as well as risks to their health and safety from exposure to hazardous materials or dangerous situations. It is important to carefully assess the potential hazards in a given workspace, taking into account factors such as the tasks that the robots will be performing, the environment in which they will be working, and the potential interactions between humans and robots.

Previous methods have been developed to separate human and robot working zones. This was done by passive means such as fences or active devices like cameras and sensors. In \cite{long2017industrial}, a system has been presented that ensures safe human--robot interaction in industrial settings. The system is designed to monitor the interactions using 3D lasers mounted on a robot. These scanners create a dynamic envelope around the robot and allow the detection of operator presence or environmental changes. Safety zones can also be set, wherein we reduce the robot speed or stop it entirely when a human enters them. Dynamically switched safety zones are used to monitor human presence and respectively adapt robot behavior, namely speed and movement status \cite{karagiannis2022adaptive}. An overhead camera system is used to identify humans entering or exiting each zone and produce warning and stop signals if necessary~\cite{scalera2020application, kim2020estimating, safeea2019minimum}.

Recent research has explored various sensing and control techniques to enhance the safety and performance of human--robot interactions. For instance, force sensing, machine learning, and wearable-technology-based safety systems have been proposed. In \cite{halme2018review, choi2022integrated, seemann2004head}, cameras and depth sensors are used to detect and track human movements and to optimize robot behavior based on safety and productivity objectives. Furthermore, emerging technologies such as \ac{AR}-assisted Deep Reinforcement Learning \cite{li2023ar}, \ac{AR}-based interaction \cite{hietanen2020ar}, and contactless collaboration with mixed reality interfaces \cite{khatib2021human} have shown great potential to improve the mutually cognitive and collaborative aspects of human--robot interactions.

Other research has focused on incorporating machine learning techniques to improve the performance of these systems \cite{oliff2020reinforcement, gao2020robust, el2020towards}. In \cite{magrini2015control, papanastasiou2019towards} force and torque sensors have been used to ensure safe human--robot interaction and collision avoidance. Wearable sensor systems are also proposed in \cite{papanastasiou2019towards, villani2020wearable, dimitropoulos2021seamless} to monitor human posture, movements, and location and to provide feedback to the robot to ensure safe interaction. A combination of visual and tactile perception interpretation is proposed in \cite{mohammadi2020mixed} to enhance detection, and thereby safety, in \ac{pHRI}.

A number of methods have been proposed to regulate the robot speed based on its position and distance to the nearest human operator \cite{kim2020estimating, byner2019dynamic}. In \cite{rosenstrauch2018human} human presence was sensed using time-of-flight sensors. It was then possible to calculate the shortest distance between the human operator and the robot, and this was used to define a scaling factor for robot speed control. Instead of reducing robot speed, a 3D-potential-field-based collision avoidance algorithm for manipulators in smart factories has been proposed in~\cite{kang2022manipulator}. The algorithm takes into account the ISO 15066 standard for safety in human--robot interaction and applies \ac{SSM} to control the speed of the manipulator based on the distance to an obstacle. In addition to the aforementioned methods, they can also be combined with filters to improve tracking results~\cite{du2018active}. The goal of the system is to ensure safety in human--robot interaction by avoiding collisions between the robot and the human. The proposed system uses the Unscented Kalman Filter (UKF) to estimate the human's position and movement, the expert system to determine appropriate avoidance strategies, and the artificial potential field method to guide the robot's movement.

Several techniques have been developed to address the limitations in the standards regarding the speed of robotic manipulators. In particular, \cite{vysocky2019motion, shin2018allowable, aivaliotis2019power} suggest checking the maximum allowable velocity of the robot and reducing its speed by a suitable factor as required. Additionally, optimization tools can be utilized to determine the maximum safe velocity for a particular path, for both scenarios where the human operator is in motion or stationary \cite{shin2018allowable}. A laser range finder was employed to detect the position of the operator, and based on the obtained data, manipulator motion planning using a Markov Model to predict the operator's position was carried out. The work presented in \cite{ferraguti2020control}~proposes a similar approach but utilized Zeroing Control Barrier Functions (ZCBFs) to enforce the velocity bounds outlined in ISO/TS~15066 \ac{PFL}. By utilizing this optimization method, it is possible to compute the required acceleration for safe motion and set bounds on velocity such that the robot can fully stop before a collision with the human operator occurs. 

The study in \cite{lucci2020combining} presents a technique for efficiently integrating \ac{PFL} and \ac{SSM} modes in collaborative robotics operations. The integration of the two modes allows for the achievement of elevated levels of productivity while still ensuring the safety of human operators. This is accomplished through the optimal scaling of the predetermined velocity while maintaining the consistency of the robot's trajectory along its path.

Further investigation is being conducted within the realm of passive compliant structures, such as collaborative soft robotics, which implement methods of dissipating impact energy to ensure safety. In the work by \cite{yoon2005safe}, a secure arm featuring passive compliant joints was devised for the purpose of creating human-friendly service robots. This design incorporated a damper and rotary spring to enhance safety. Similarly, ref.~\cite{park2008safe} introduced a secure linkage mechanism that consists of linear springs and a double-slider arrangement, enabling the mechanism to possess adaptable stiffness and ensure safety in scenarios involving collisions with humans. Likewise, in \cite{wolf2008new}, a novel robotic joint endowed with variable stiffness was proposed. This design harnessed mechanical compliance to store potential energy, resulting in a more streamlined and lightweight configuration compared to analogous models. In the work by \cite{seriani2018development}, a comprehensive framework for modeling structures with multiple degrees of freedom (\emph{n}-\ac{DoF}) with a specific emphasis on robots was presented. The proposed analytical model was validated through impact tests conducted on a designated structure, demonstrating effective passive collision~protection.

The safety mechanisms utilized in commercial collaborative robotic systems have undergone substantial evolution. Numerous robots now exploit sophisticated functionalities like lightweight structures, collision detection mechanisms, and minimized points susceptible to accidental pinching. Despite these notable advancements, it remains imperative to implement safety protocols that encompass all robot elements such as the gripper, end-effector, and proximate equipment within the collaborative workspace. Manufacturers also incorporate safety apparatuses to ensure the safeguarding of this shared workspace. Among these measures are safety area scanners and mats, safety light curtains, and safety switches \cite{martinetti2021redefining}. Furthermore, in situations where potential hazards are identified or the operation bears the potential to generate risks, operators have the option to engage the ``deadman'' switch, a control that automatically stops the system should the user fail to sustain pressure.

In the context of research in this domain, where most endeavors have concentrated on implementing the fundamental speed and positional constraints established by standards, a noteworthy gap persists concerning methodologies that can dynamically recalibrate robot parameters to reconfigure these constraints for heightened operational efficacy. To address this research gap, this paper proposes a novel approach that enhances operation speed while ensuring safety within human--robot collaborative situations. The control scheme utilizes a variable impedance algorithm for robotic manipulators, following the ISO/TS~15066 standard. 

In this paper, our main contributions are to:

\begin{enumerate}
\item Propose an impedance-controller-based approach to lower the effective mass of a robot, thus enabling higher operational speeds while adhering to safety restrictions of ISO/TS 15066 \ac{PFL} mode;
\item Present a formulation for setting the effective mass parameters as a function of the robot movement direction;
\item Analyze the effects of parameter selection on the aforementioned controller;
\item Perform a comparative assessment of the proposed controller and conventional control methods, including \ac{PD} and \ac{CTM};
\item Demonstrate the efficacy of the method on two distinct robotic platforms: a generic 3R manipulator and the Franka Erika Panda.

\end{enumerate}

The structure of this paper is as follows. \refsec{sec:2} describes the power and force limiting mode of the ISO/TS~15066 standard. The necessary steps and parameters for calculating the maximum permissible velocity are outlined. In \refsec{sec:3}, the variable impedance controller in both joint-space and workspace is formulated, along with a method for determining the reduced effective mass parameters. In \refsec{sec:4} the path-planning algorithm and controller block diagram for real-time implementation of the proposed method is stated. The simulation results of our proposed method on the 3R manipulator and Franka Erika Panda robots are presented in \refsec{sec:5}. Finally, \refsec{sec:6} concludes the work and provides recommendations for future research in this field.

\section{Safety in ISO/TS~15066 Power and Force Limiting Mode}
\label{sec:2}

The development of safety standards for the interaction between humans and robots is of utmost importance. To address this issue, the \ac{ISO} has established two crucial standards: ISO~10218-1 and ISO~10218-2, entitled ``Robots and Robotic Devices---Safety Requirements for Industrial Robots''. These standards provide fundamental safety requirements for the deployment of robots in industrial settings. In addition to the aforementioned standards, ISO/TS~15066 serves as a supplement and outlines the guidelines for the integration of collaborative robots into the work environment.

The ISO/TS~15066 technical specification is a vital document for ensuring safe collaborative robot operation in shared workspaces. The increasing trend towards human--robot collaboration has made it necessary to pay close attention to the safety of these systems, and the ISO/TS~15066 provides a comprehensive approach to achieve this goal. The document outlines guidelines for the safe integration of collaborative robots into the work environment and focuses on the importance of maintaining the integrity of the safety-related control system. This is particularly crucial in collaborative robot operations wherein the robot system and human workers share the same workspace, and process parameters such as speed and force are being controlled. The guidelines provided by ISO/TS~15066 are essential for ensuring safe and efficient collaboration between humans and robots, and they serve as a reference for the development of industry best practices. 

The standard outlines four distinct modes of collaborative operation: (1) safety-rated monitored stop; (2) hand guiding; (3) speed and separation monitoring; and (4) power and force limiting. The present study explores the utilization of two operating modes in robotic systems, namely hand guiding and power and force limiting. These modes allow for physical interaction between the robot and an operator. The choice of these two modes is motivated by the intent to employ impedance control in the analysis, as impedance control is deemed suitable for the interaction of the robot with the environment---particularly, with a human in this case. The focus of the research is to investigate control mechanisms in the context of \ac{PFL} mode.

In the present operational mode, the potential for physical interaction between the robot and the operator can occur as a result of intentional or unintentional contact. The paramount consideration in such scenarios is the reduction of risks and the provision of a safe working environment. This objective is achieved by operating the robot at a velocity below a specified threshold value. The determination of this threshold velocity is of utmost importance, as it governs the maximum velocity at which the robotic system may move. This maximum velocity can be calculated using a formula that takes into account various safety parameters and constraints
\begin{equation}
v_\text{rel, max} = \frac{F_{\text{max}}}{\sqrt{\mu k}}, 
\label{eq:1}
\end{equation} 
where $v_\text{rel, max}$ is the maximum relative safe velocity, $F_\text{max}$ is the maximum allowable force that can be imposed on a specific body part, $k$ is the effective spring constant of the relevant body segment, and $\mu$ is the reduced mass of the two body system, given by
\begin{equation}
\label{eq:reduced}
\mu = (\frac{1}{m_\text{H}} + \frac{1}{m_\text{R}})^{-1},
\end{equation} 
where $m_\text{H}$ is the effective mass of the human body segment and $m_R$ is the effective mass of the robotic system. The parameter $m_\text{R}$ is a function of robot posture and position and can be performed using various methods, including conservative formulas and more precise methods, which will be addressed in \refsec{sec:4}. The velocity threshold is impacted by various factors, including human and robot mass characteristics. To account for human factors, a mechanism must be established for detecting which body segments could potentially interact with the robot's end-effector or other mobile components~\cite{ISOTS15066}. Upon identification of all relevant body segments, the parameters $F_\text{max}$ and $k$ can be determined.

\begin{table}[!t]
\caption{Parameters of the human body model given in ISO/TS 15066 \cite{ISOTS15066}.}
\centering
\begin{tabularx}{\textwidth}{@{}X>{\centering\arraybackslash}X>{\centering\arraybackslash}X>{\centering\arraybackslash}X@{}}
\toprule
\textbf{Body Region} & \textbf{Maximum}& \textbf{Effective Spring}& \textbf{Effective Mass} (kg)\\
 & \textbf{Permissible Force} (N)& \textbf{Constant K} (N/mm)& \\ 
\midrule
Skull and forehead & 130 & 150 & 4.4 \\
Face & 65 & 75 & 4.4 \\
Neck & 150 & 50 & 1.2 \\
Back and shoulders & 210 & 35 & 40 \\
Chest & 140 & 25 & 40 \\
Abdomen & 110 & 10 & 40 \\
Pelvis & 180 & 25 & 40 \\
Upper arms and elbow & 150 & 30 & 3 \\
joints & & &  \\
Lower arms and wrist & 160 & 40 & 2 \\
joints &  &  &  \\
Hands and fingers & 140 & 75 & 0.6 \\
Thighs and knees & 220 & 50 & 75 \\
Lower legs & 130 & 60 & 75 \\
\bottomrule
\end{tabularx}
\label{tab:1}
\end{table}

The correlation between biomechanical constraints and the transfer of energy during contact events must be thoroughly considered. As demonstrated by \refeq{eq:1}, parameters such as the maximum permissible force, the effective mass, and the spring constant of the affected body part are utilized to compute the maximum velocity limit. The body model and its constituent parts are presented in the relevant standard. The body is initially divided into two sections, the front and rear, which are further sub-divided into smaller sections as depicted in \reftab{tab:1}. As a general principle, it can be observed that smaller and more delicate body parts exhibit lower tolerance to impact forces, thereby resulting in lower threshold values.

\section{Robot Effective Mass Calculation}
\label{sec:3}

The determination of the maximum permissible velocity necessitates the calculation of the effective mass of the robotic system. This value is dependent on various factors, including the robot's posture, motion, and payload. This section presents and compares three methods for determining the effective mass. A simulation-based comparison of these methods is presented afterwards.

\subsection{Effective Mass from ISO/TS 15066}

\begin{figure}[t!]
\centering
\subfloat{\includegraphics[width=0.4\linewidth]{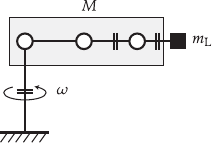}}
\caption{Schematic of the effective mass for the robot moving parts.}
\label{fig:3}
\end{figure}

The method proposed in the ISO/TS~15066 standard involves the calculation of the robot's effective mass. This calculation is performed by taking half of the sum of the masses of all the moving parts of the robot and its payload, where these parts are shown in \reffig{fig:3}. This formula is known to produce conservative results, leading to a higher calculated effective mass value. As the effective mass appears in the denominator of the equation determining the maximum velocity, this results in a corresponding reduction in the maximum permissible velocity
\begin{equation}
m_\text{R} = \frac{M}{2} + m_\text{L},
\end{equation} 
where $m_L$ is the effective payload of the system, including tooling and workpiece, and $M$ is the total mass of the moving parts of the robotic system.

\subsection{Effective Mass from Operational Space Inertia Matrix}

The effective mass of the robot, denoted as $m_R$, at any given configuration $q$ along the direction of $u$ can be expressed as follows \cite{khatib1995inertial}
\begin{equation}
\label{khatib}
m_\text{R}^{-1} = u^T J(q) M^{-1} J(q)^T u,
\end{equation}
where $q \in {\mathbb{R}^n}$ represents the vector of joint angles of the robot in a given position, $M(q)$ represents the inertia matrix, and $J(q)$ represents the Jacobian matrix of the manipulator.

The equation can also be written based on the workspace inertia matrix $\bar M$ as
\begin{equation}
\label{khatibws}
m_\text{R}^{-1} = u^T {\bar M^{-1}} u.
\end{equation}

\subsection{Variable Effective Mass with Impedance Control} 

The mechanical impedance refers to the ratio of force output to motion input; all physical objects exhibit this property, and they can be modeled as spring--mass--damper systems. Impedance control enables regulation of the relationship between force and position, velocity, and acceleration. By utilizing impedance control, it is possible to design a control law that provides a desired, specified impedance in relation to external forces~\cite{hogan1984impedance, hogan1985impedance}. This method involves tuning of the mass--damper--spring coefficients. Using this method, by reducing the effective mass as described in \refeq{eq:1} from the \ac{PFL} mode defined in \mbox{ISO/TS 15066}, it becomes possible to operate the robot at increased velocities, which, in turn, results in lead-time reduction.

Following this section are the steps required to control a robotic system based on the proposed impedance controller. In this method, we treat the environment and the operator as an admittance; thus, the robot should act as an impedance. The control law can modulate the impedance of the manipulator according to the task. Since the formula of maximum permissible velocity is a function of the robot's mass value alone, we focus on tuning this parameter and leave stiffness and damping unchanged.

The dynamic model of a robotic manipulator in joint space coordinates has the form
\begin{equation}
\label{eq:robotdyn}
M(q) \ddot q + C(q,\dot q) \dot q + G(q) = u + J^T F_{ext},
\end{equation}
where $q \in {\mathbb{R}^n}$ represents the vector of joint angles of the robot in a given position, $M(q)$ is the inertia matrix, $C(q,\dot q) \dot q$ is the vector of Coriolis/centrifugal torques, $G(q)$ represents potentials such as gravity, $u$ is the vector of controller torque values, $J(q)$ is the robot Jacobian, and $u_{ext} = J^T F_{ext}$ is the joint torque resulting from external force and torque $F_{ext}$ applied at the end-effector.

In the context of robot manipulation, the controller is responsible for exerting a force to effect object movement. This force can be represented by a virtual spring and damper system. When the robot interacts with a human operator who attempts to move it, the control law is designed to generate a force that opposes the motion in a manner to simulate a spring--damper system. By choosing the control input $u$ in the dynamic equation of a robot as modeled in \refeq{eq:robotdyn} with
\begin{equation}
\label{eq:u11}
u = M(q) y + C(q,\dot q) + G(q),
\end{equation}
it simplifies to
\begin{equation}
\label{eq:u12}
\ddot q = y + M^{-1} (q) J^T (q) F_\text{ext}.
\end{equation}
Due to the contact forces, a nonlinear coupling term is present in \refeq{eq:u12}. If $y$ is taken to be
\begin{equation}
\label{eq:u13}
y = J^{-1} (q) M_d^{-1} \left( M_d \ddot r_d - K_d \dot e _r - K_p e_r - M_d \dot J (q, \dot q) \dot q\right), 
\end{equation}
where $e_r = r - r_d$ is the vector of displacements in workspace coordinates, $M_d$ is a diagonal positive definite matrix, and $K_d$, $K_p$ are positive definite matrices, then \refeq{eq:u12} equates to
\begin{equation}
\label{eq:u14}
{M_d}{{\ddot e}_r} + {K_d}{{\dot e}_r} + {K_p}{e_r} = {M_d}J(q){M^{ - 1}}(q){J^T}(q){F_\text{ext}},
\end{equation}
which can be simplified as
\begin{equation}
{M_d}{{\ddot e}_r} + {K_d}{{\dot e}_r} + {K_p}{e_r} = {M_d}{{\bar M}^{ - 1}}(q){F_\text{ext}}.
\end{equation}

A generalized mechanical impedance is established to relate the vector of forces ${M_d}{{\bar M}^{ - 1}}(q){F_{ext}}$ to the vector of displacements $e_r = r - r_d$. The dynamic behavior within the workspace can be specified by modifying this mechanical impedance. The inclusion of the term ${{\bar M}^{ - 1}}(q)$ results in a coupled system. In order to maintain linearity and decoupling during interaction with the environment, it is imperative to consider the contact forces during control. If the error-free force measurements are available, the input could be selected as
\begin{equation}
\label{eq:u15}
u = M(q) y + C(q,\dot q) + G(q) - J^T (q) F_\text{ext},
\end{equation}
with 
\begin{equation}
\label{eq:u16}
y = J^{-1} (q) M_d^{-1} \left( M_d \ddot r_d - K_d \dot e _r - K_p e_r - M_d \dot J (q, \dot q) \dot q + F_\text{ext} \right). 
\end{equation}
Replacing \refeq{eq:u15} and \refeq{eq:u16} in \refeq{eq:robotdyn} results in
\begin{equation}
\label{eq:u17}
{M_d}{{\ddot e}_r} + {K_d}{{\dot e}_r} + {K_p}{e_r} = {F_\text{ext}}.
\end{equation}
The expression $J^T(q) F_\text{ext}$ in Equation \refeq{eq:u15} serves the purpose of compensating for contact forces and conferring infinite stiffness upon the robot. However, the introduction of the term $J^{-1}(q)M_d^{-1} F_\text{ext}$ allows for the imposition of a compliant behavior in the robot, which can be characterized by a linear impedance with respect to $F_\text{ext}$.

If the dimensions of the configuration space $Q$ are larger than the dimensions of the workspace, then the Jacobian is not square and its inverse is not defined. In such cases, the pseudo-inverse is used
\begin{equation}
\left\{
{\begin{array}{ll}
J^+ = J^T(q) \left( J(q) J^T(q) \right) ^{-1} \text{, if $J$ is $m \times n$ with $m<n$}, \\
J^+ = \left( J(q) J^T(q) \right) ^{-1} J^T(q) \text{, if $J$ is $m \times n$ with $m>n$}. 
\end{array}} \right.
\end{equation}

A limitation when setting the desired inertia matrix $M_d$ for impedance control is that it is difficult to accurately estimate the inertia matrix of a system. This is especially true when dealing with complex robotic systems, as it can be difficult to accurately measure the inertial properties of each component. Additionally, if the desired inertia matrix is set too high, it can lead to an overly stiff system, which may be difficult to control and could lead to instability. On the other hand, if the desired inertia matrix is set too low, then the system may become too compliant and it could lead to excessive vibrations or oscillations. Therefore, it is important to carefully consider the desired inertia matrix when setting up an impedance control system in order to ensure optimal performance.

Given the aforementioned limitations, we propose a method to derive the desired inertia matrix, denoted by $M_d$, in the workspace coordinates. As previously mentioned, the mass of the object along the direction of the unit vector $u$ is computed using \refeq{khatibws}. To account for the effects of the environment and to improve the robot's responsiveness, a reduction factor $\lambda$ is introduced to effectively reduce the mass. Specifically, $\lambda$ is a scalar value in the range of $0\leqslant\lambda<1$, which is applied to the calculated mass to obtain the reduced effective~mass
\begin{equation}
{m_R}^\prime  = \lambda {m_R}.
\end{equation}
Substituting values into \refeq{khatibws} results in
\begin{equation}
u^T M_d^{-1} u = u^T \frac{\bar M^{-1}}{\lambda} u.
\end{equation}
The equation can be expressed in scalar values on both sides, thus allowing for a rewritten form
\begin{equation}
\label{summation}
\sum ({u u^T \odot M_d^{-1}}) = \sum ({u u^T \odot \frac{\bar M^{-1}}{\lambda}}),
\end{equation}
where $\odot$ denotes element-wise matrix multiplication, while $\sum$ represents the summation of all the elements in the resulting matrix. Assuming that ${\bar M^{-1}}$ and $M_d^{-1}$ are $n \times n$ matrices and that $u$ is an $n \times 1$ vector, \ref{summation} can be expressed using indices as follows
\begin{equation}
\label{cond}
\sum\limits_{i = 1}^n {\sum\limits_{j = 1}^n {{u_i}{u_j}{M_d}{{^{ - 1}}_{i,j}}}} = \frac{1}{\lambda} \sum\limits_{i = 1}^n {\sum\limits_{j = 1}^n {{u_i}{u_j}{\bar M}{{^{ - 1}}_{i,j}}}}. 
\end{equation}
The desired mass matrix $M_d$ is chosen to be a diagonal positive definite matrix. A suitable form for this matrix is as follows
\begin{equation}
{M_d}^{ - 1} = \left( {\begin{array}{*{20}{c}}
  {{\gamma _1}}&0&0& \ldots &0 \\ 
  0&{{\gamma _2}}&0& \ldots &0 \\ 
  0&0&{{\gamma _3}}& \ldots &0 \\ 
   \vdots & \vdots & \vdots &{}& \vdots \\ 
  0&0&0& \ldots &{{\gamma _n}} 
\end{array}} \right).
\end{equation}
A selection of diagonal elements $\gamma_i$, while satisfying the condition \refeq{cond}, is as follows
\begin{equation}
{\gamma _i} = \frac{1}{{\lambda {u_i}}}\sum\limits_{j = 1}^n {{u_j}{\bar M}{{^{ - 1}}_{i,j}}}.
\end{equation}
Consequently the new effective mass ${m_u}^\prime$ in the direction $u$ equates to
\begin{equation}
{m_R}^\prime = \lambda m_R = \frac{1}{u^T M_d^{-1} u}.
\end{equation}

\section{Control Design and Path Planning}
\label{sec:4}

To enable a comprehensive comparison, two additional controllers, namely \ac{PD} control and \ac{CTM} have been employed alongside the impedance controller. \ac{PD} and \ac{CTM} are designed in workspace coordinates with parameters similar to those of the impedance controller. For path planning, the maximum speed of the robot is considered in order to ensure safety in the application of these controllers.

The \ac{PD} input law for a set of mechanical equations \refeq{eq:u11} is
\begin{equation}
u = -J^T(q) K_p e_r - J^T(q) K_d J(q) \dot q + G(q),
\end{equation}
with $e_r = r - r_d$, where $r_d$ is a constant vector of desired workspace coordinates. Parameters $K_d$ and $K_p$ are both design parameters that are positive definite matrices. With this choice, the control law renders the equilibrium asymptotically stable. The resulting closed-loop dynamis \refeq{eq:robotdyn} for the \ac{PD} controller becomes
\begin{equation}
M(q) \ddot q + C(q,\dot q) \dot q = -J^T(q) K_p e_r - J^T(q) K_d J(q) \dot q.
\end{equation}
The control law of the \ac{CTM} controller in workspace coordinates is taken to be
\begin{equation}
\begin{gathered}
u = M(q) y + C(q, \dot q) \dot q + G(q), \\
y = J^{-1}(q) \left( \ddot r_d - K_d \dot e_r - K_p e_r - \dot J(q, \dot q) \dot q \right).
\end{gathered}
\end{equation}
The resulting closed-loop dynamic for the \ac{CTM} controller is
\begin{equation}
{{\ddot e}_r} + {K_d}{{\dot e}_r} + {K_p}{e_r} = 0.
\end{equation}

\subsection{Path Planning} 

The velocity constraint $v_\text{rel}$ specified in the standard is restricted to the direction between the operator and the robotic manipulator, and its calculation must be performed in real-time due to the constantly changing direction. To simplify analysis without loss of generality, the human operator is assumed to be stationary in the simulations. The maximum velocity calculated according to the constraint is then projected onto the desired path direction, representing the desired velocity for the robotic manipulator.

\begin{figure}[t!]
\centering
\subfloat{\includegraphics[width=0.55\linewidth]{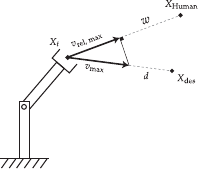}}
\caption{Schematic of the velocity vector $v_\text{max}$ computed in relation to the human $w$---oriented in the direction of their mutual correspondence---and its relationship with the maximum relative velocity vector $v_\text{rel,max}$, which is oriented in the direction of the intended motion $d$.}
\label{fig:vectors}
\end{figure}

The maximum velocity that the robot can attain $|v_\text{rel,max}|$ can be computed based on the human--robot mass, the maximum allowable force, and the effective spring constant. This scalar quantity is oriented along the direction vector $w$, which lies between the end-effector of the robot and the human. In this paper, we assume that the distance between the robot and the human is measured from the robot's end-effector. However, if needed, this calculation can be done from any other point on the robot without loss of generality. As a result, the formula for calculating the maximum relative velocity vector $v_\text{rel,max}$ considering both human and end-effector coordinates is given by
\begin{equation}
v_\text{rel,max} = |v_\text{rel,max}| \times \frac{X_{Human} - X_t}{||X_{Human} - X_t||}.
\end{equation}
To determine the maximum velocity in the desired direction $v_\text{max}$, this value is projected onto the vector $d$, representing the desired path
\begin{equation}
v_\text{max} = v_\text{rel,max}\cdot \frac{X_{des} - X_t}{||X_{des} - X_t||},
\end{equation}
where ``$\cdot$'' is vector dot product. The different vectors are shown in the diagram in \reffig{fig:vectors}. Once the operational speed of the robot has been determined, the desired position of the robot at a time step of $dt$ can be computed assuming no acceleration as follows
\begin{equation}
X_{t+dt} = X_{t} + v_\text{max} \times dt.
\end{equation}

\subsection{Controller Block Diagram}

\begin{figure}[t!]
\centering
\subfloat{\includegraphics[width=1 \linewidth]{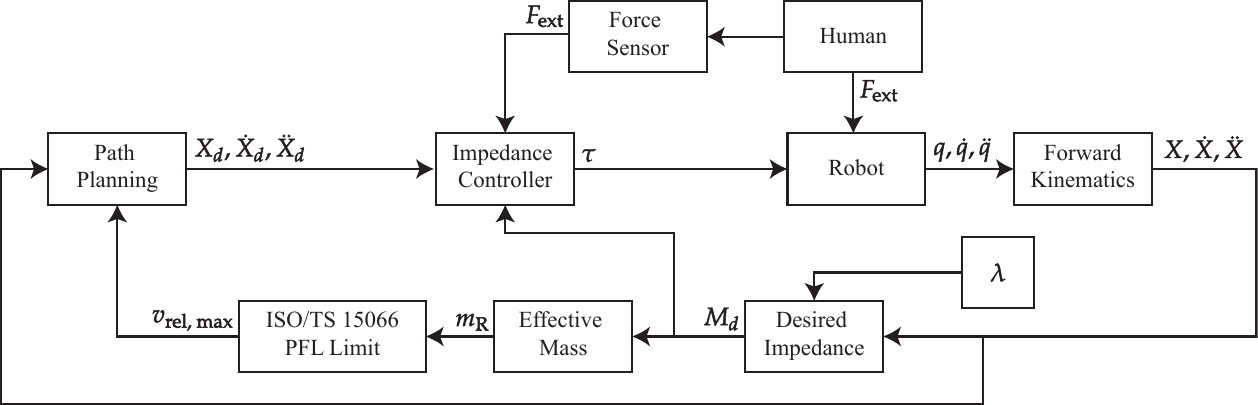}}
\caption{Block diagram of the proposed variable impedance controller.}
\label{fig:block}
\end{figure}

For simulation of the variable impedance algorithm, the block diagram shown in \reffig{fig:block} is setup. It has components for online path planning and real-time calculation of maximum safe velocity. 
To investigate errors that arise during the simulation of robots, it is necessary to establish a metric for measuring such errors. Several metrics such as concurrent motion, human idle time, total task time, and concurrent activity in the robot workspace have been proposed \cite{dragan2015effects, hoffman2019evaluating, scalera2022enhancing}. The metric employed in this paper is the absolute tracking error in Cartesian coordinates across the entirety of the simulated trajectory, as expressed by the following equation
\begin{equation}
{{\bar e}_i} = \frac{1}{N}{\sum\limits_{n = 1}^N {\left| {{{e}_i}(n)} \right|} },
\end{equation}
where $N$ denotes the total number of observations obtained for each error variation. The settling time (total task time) is also calculated for the experiments. 

\section{Simulation Results}
\label{sec:5}

In order to assess the efficacy of the proposed controller, a series of simulations have been conducted on two distinct robotic manipulators. The first is a 3R robot operating within a two-dimensional workspace. The second is a Franka Emika Panda robot---a 7-DoF industrial cooperative robot---simulated within Robotics Toolbox for Python (v1.1.0) \cite{rtb} simulator environment. The simulation results indicate that the proposed control scheme is capable of enabling these manipulators to operate at significantly higher speeds than alternative control strategies. It is shown that this approach is effective in limiting contact forces so that they remain below established safety thresholds. This feature is of critical importance in preventing injuries and ensuring that the operation of the manipulators complies with all relevant safety standards.

\subsection{The 3R Robot in a 2D Workspace}

\begin{figure}[t!]
\centering
\subfloat{\includegraphics[width=0.35\linewidth]{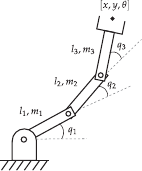}}
\caption{Schematic of the simulated 3R robot in 2D workspace.}
\label{fig:rrr}
\end{figure}

The performance of the proposed impedance controller is initially investigated on a 3R robotic manipulator operating in a 2D workspace. A schematic of the robot is shown in \reffig{fig:rrr}. Given that the manipulator is functioning within a Cartesian 2D coordinate system, the joint angle variables are represented as a three-element vector $Q = [q_1, q_2, q_3]^T \in {\mathbb{R}^3}$, while its workspace variables are defined as a three-dimensional vector $r = [x,y,\theta]^T \in {\mathbb{R}^3}$. The translational Jacobian matrix of the 3R robot used is given as
\begin{equation}
J(q) = \begin{bmatrix}
    -l_1 s_1 - l_2 s_{12} - \frac{1}{2} l_3 s_{123} & -l_2 s_{12} - \frac{1}{2} l_3 s_{123} & -\frac{1}{2} l_3 s_{123} \\
    l_1 c_1 + l_2 c_{12} + \frac{1}{2} l_3 c_{123} & l_2 c_{12} + \frac{1}{2} l_3 c_{123} & \frac{1}{2} l_3 c_{123} \\
    1 & 1 & 1 \\
\end{bmatrix}
\label{eq:jacobian}
\end{equation}
where $s_i~(c_i)$ denotes $sin(q_i)~(cos(q_i))$, and $s_{ij}~(c_{ij})$ denotes $sin(q_i+q_j)~(cos(q_i+q_j))$. The inertia tensor is given by
\begin{equation}
M(q) = \begin{bmatrix}
I_1 + I_2 + I_3 + 2I_{12} + 2I_{23} + 2I_{13} & I_2 + I_3 + I_{23} + \frac{1}{2} I_{13} & I_3 + \frac{1}{2} I_{23} + \frac{1}{2} I_{13} \\
I_2 + I_3 + I_{23} & I_2 + I_3 + I_{23} & I_3 + \frac{1}{2} I_{23} \\
I_3 + \frac{1}{2} I_{23} + \frac{1}{2} I_{13} & I_3 + \frac{1}{2} I_{23} & I_3 \\
\end{bmatrix}, 
\end{equation}
where
\begin{equation}
{I_1} = \frac{1}{3}{m_1}l_1^2,~{I_2} = \frac{1}{3}{m_2}l_2^2,~{I_3} = \frac{1}{3}{m_3}l_3^2,~{I_{12}} = {m_2}{l_1}{l_2}{c_2},~{I_{23}} = {m_3}{l_2}{l_3}{c_3},~{I_{13}} = {m_3}{l_1}{l_3}{c_{23}}.
\label{eq:inertiatensor}
\end{equation}
The physical parameters ($l_i$ and $m_i$) of this robot are also reported in \reftab{tab:rrrparams}. 
\begin{table}[!t]
\caption{Physical parameters of 3R planar robot}
\centering
\begin{tabularx}{\textwidth}{@{} >{\hsize=1.2\hsize}X >{\centering\arraybackslash}>{\hsize=0.6\hsize}X >{\centering\arraybackslash}>{\hsize=0.6\hsize}X >{\centering\arraybackslash}>{\hsize=0.6\hsize}X @{}}
\toprule
\textbf{Parameter} & \textbf{Link 1}& \textbf{Link 2}& \textbf{Link 3}\\
\midrule
Mass (kg) & 8 & 5 & 5 \\
Length (m) & 2 & 2 & 2 \\
Moment of inertia (kgm$^2$) & 10.66 & 6.66 & 6.66 \\
\bottomrule
\end{tabularx}
\label{tab:rrrparams}
\end{table}

The manipulator is simulated in MATLAB/Simulink software, which allows for precise analysis of its behavior. With a high degree of precision, the manipulator can accurately position and control objects in its workspace. This flexibility and range of motion make it well-suited for complex tasks. The variable impedance controller is used to dynamically adjust the robot's end-effector impedance in response to changes in the environment. The controller's effectiveness is demonstrated through experimental results, which show improved performance compared to other controllers.

In the simulation, it is assumed that a human operator is situated at the stationary position of $X_\text{Human} = [7, 2.5]^T$ within the workspace. The abdominal section of the human body is placed in close proximity to the end effector of the robot. The required parameters of the abdomen body part are referenced in \reftab{tab:1}. This is done for simulation purposes, as in practice it is necessary to use a suitable system to accurately determine the precise location and specific body part of the human operator. The robot's base is positioned at the origin and the manipulator links start at angles of $Q_\text{t=0} = [\frac{3}{4}\pi, -\frac{1}{2}\pi, -\frac{1}{2}\pi]^T$. As a result, the robot's end effector is initially located at the position $X_\text{robot, t=0} = [\sqrt{2}, \sqrt{2},-\frac{1}{4}\pi]^T$. 

Each controller is required to command the robot in a way that results in its motion along a straight path from the initial location to a designated end effector position, $X_\text{robot} = [5, \sqrt{2},-\frac{1}{12}\pi]$. It should be noted that the determinant of the Jacobian matrix is greater than or equal to 3.95 along all the intermediary configurations, which implies that the path is valid and avoids intersecting any singular points of the robot. Each controller employs the path planning equations introduced in the preceding section, and they ensure that the robot is operating at its maximum allowable speed, which is determined by the ISO/TS~15066 \ac{PFL} mode.

\begin{figure}[t!]
\centering
\subfloat[\centering PD Controller]{\includegraphics[width=0.45\linewidth]{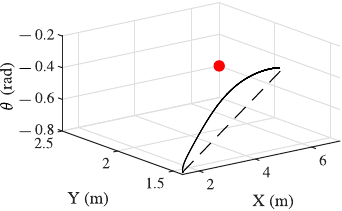}}
\subfloat[\centering CTM Controller]{\includegraphics[width=0.45\linewidth]{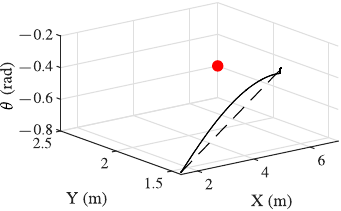}}
\
\subfloat[\centering $\text{IMP}_1$ Controller]{\includegraphics[width=0.45\linewidth]{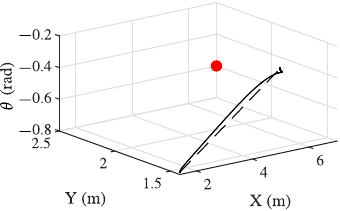}}
\subfloat[\centering $\text{IMP}_2$ Controller]{\includegraphics[width=0.45\linewidth]{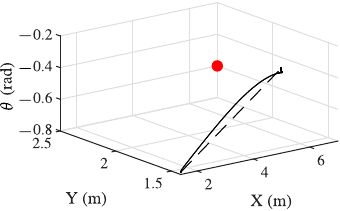}}
\caption{Trajectory of the 3R robot end-effector in workspace coordinates (dashed: desired; solid: actual). The red dot depicts the position of a generic obstacle in the workspace that is stationary during robot movement. The obstacle is taken to be the abdominal section of a human operator for simulation purposes.}
\label{fig:rrrtraj}
\end{figure}

\begin{figure}[t!]
\centering
\subfloat{\includegraphics[width=0.9\linewidth]{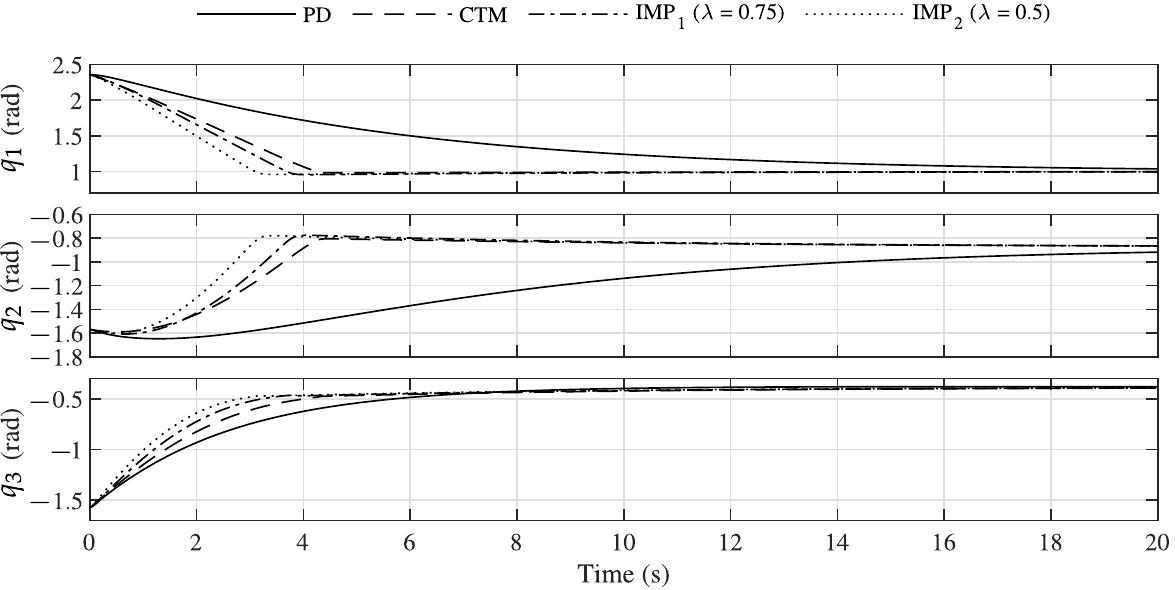}}
\caption{Instantaneous trajectory of the 3R robot end-effector in joint space coordinates.}
\label{fig:rrrjoint}
\end{figure}

\begin{figure}[t!]
\centering
\subfloat{\includegraphics[width=0.9\linewidth]{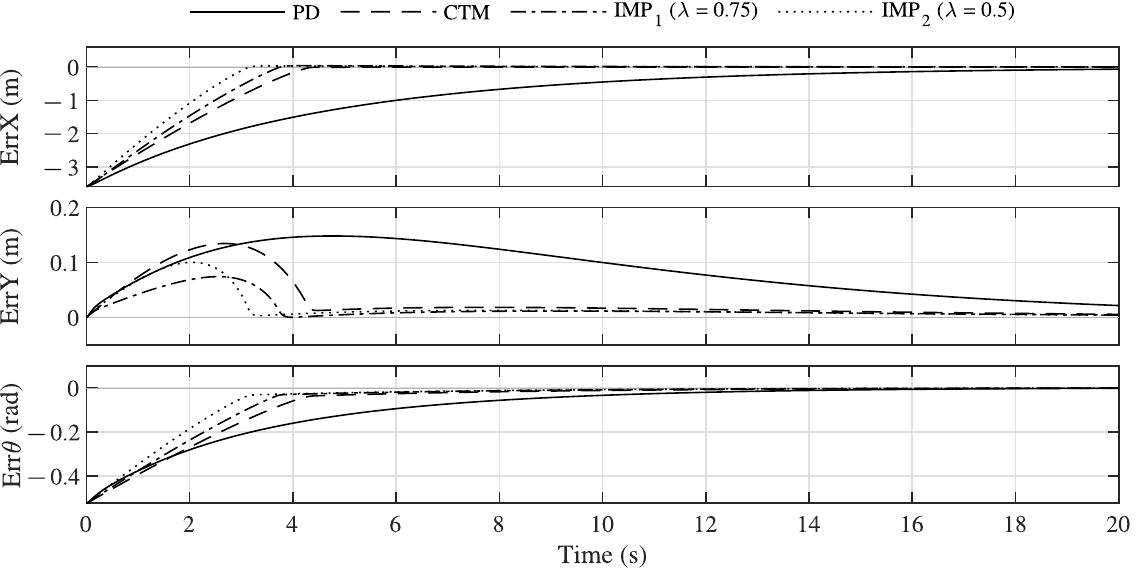}}
\caption{Instantaneous tracking errors of workspace coordinates of 3R robot.}
\label{fig:rrrerr}
\end{figure}

To maintain consistency between each scenario, the controller parameters $K_p$ and $K_d$ are held constant, with values of $K_p = 20$ and $K_d = 100$. The selection of these values was undertaken through an empirical process involving an assessment of system performance across a range of $K_p$ values spanning from 10 to 50 in increments of 5, as well as values ranging from 50 to 200 with increments of 50 for $K_d$. The selection of a high value for $K_d$ is a deliberate choice intended to guarantee that the robot operates at the maximum velocity permitted by the established standard, while the choice of $K_p$ is to maintain a low overall position error. The performance of the impedance controller is assessed via two distinct simulations---in which $\lambda$ takes on the values of 0.75 ($\text{IMP}_1$) and 0.5 ($\text{IMP}_2$)---in conjunction with the \ac{CTM} and \ac{PD} controllers. Each of these controllers is applied to the 3R robot, and the ensuing Cartesian position of the end-effector is depicted in \reffig{fig:rrrtraj}. Additionally both instantaneous trajectory of the robot in joint-space coordinates and tracking error in workspace coordinates are shown in \reffig{fig:rrrjoint} and \reffig{fig:rrrerr} respectively. 
 
The completion of the task is determined by the x component of the end-effector position reaching $95\%$ of the final desired value. The duration taken to achieve this is referred to as the settling time. The controllers' settling times are $12.95$, $3.86$, $3.39$, and $2.84$ s for \ac{PD}, \ac{CTM}, $\text{IMP}_1$ ($\lambda = 0.75$), and $\text{IMP}_2$ ($\lambda = 0.5$), respectively. Based on the results, $\text{IMP}_1$ demonstrated a $74\%$ improvement in settling time over \ac{PD} and a $12\%$ improvement over \ac{CTM} controllers. Similarly, $\text{IMP}_2$ showed a greater improvement, with a reduction in settling time of $78\%$ compared to the \ac{PD} and $26\%$ compared to the \ac{CTM} controllers. These results suggest that by carefully adjusting the parameter $\lambda$, better performance and faster speeds can be attained.

Upon assessment of various controllers, the outcomes are presented in \reftab{tab:3rreport}. The variable impedance controllers produced within the confines of this investigation are observed to exhibit substantial progress in minimizing tracking error values across all three task-space coordinates.

\begin{table}[t!]
\caption{Comparison between the performance of simulated controllers of 3R robot.}
\centering
\begin{tabularx}{\textwidth}{@{} X >{\centering\arraybackslash}X>{\centering\arraybackslash}X>{\centering\arraybackslash}X>{\centering\arraybackslash}X @{}}
\toprule
\textbf{} & \textbf{Settling}& \multicolumn{3}{c}{\textbf{Average tracking error in task-space}} \\
\cmidrule(lr){3-5}
 & \textbf{time} (s) & $\mathrm{\mathbf{{\bar e}_x}}$ (m)& $\mathrm{\mathbf{{\bar e}_y}}$ (m) & $\mathrm{\mathbf{{\bar e}}}_{\boldsymbol{\theta}}$ (rad) \\
\midrule
\textbf{PD} & 12.95 & 0.8844 & 0.0855 & 0.0967 \\
\textbf{CTM} & 3.86 & 0.3903 & 0.0301 & 0.0707 \\
$\mathrm{\mathbf{IMP_1}}$ & 3.39 & 0.3593 & 0.0158 & 0.0618 \\
$\mathrm{\mathbf{IMP_2}}$ & 2.84 & 0.3046 & 0.0176 & 0.0541 \\
\bottomrule
\end{tabularx}
\label{tab:3rreport}
\end{table}

\begin{figure}[t!]
\centering
\subfloat[\centering PD Controller]{\includegraphics[width=0.45\linewidth]{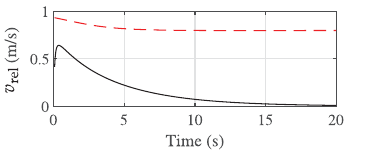}} \quad
\subfloat[\centering CTM Controller]{\includegraphics[width=0.45\linewidth]{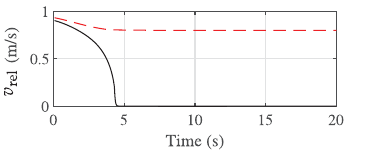}}
\
\subfloat[\centering $\text{IMP}_1$ Controller]{\includegraphics[width=0.45\linewidth]{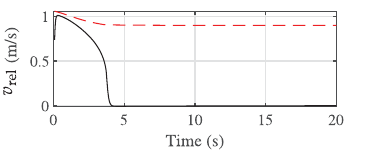}} \quad
\subfloat[\centering $\text{IMP}_2$ Controller]{\includegraphics[width=0.45\linewidth]{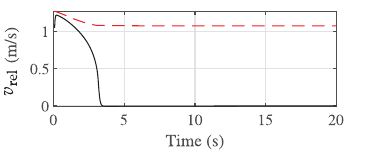}}
\caption{Instantaneous controller relative velocity of 3R robot (dashed red: ISO/TS 15066 PFL maximum velocity; solid: actual velocity).}
\label{fig:rrrspeed}
\end{figure}

It is expected that each controller will cause the robot's velocity $v_\text{rel}$ relative to the human operator in the workspace to attain the highest allowable velocity based on the effective mass of the manipulator robot and the direction of motion. The maximum velocity limit is inherently influenced by factors such as the effective mass and direction of velocity of the robot, which vary throughout the trajectory. The relative velocity of the robot and its maximum velocity for all the controllers compared are presented in \reffig{fig:rrrspeed}. This demonstrates the effectiveness of each control approach in governing the robot's motion along its intended trajectory. A controller's ability to maintain the maximum speed is enhanced if the velocity approaches the maximum allowable speed.

As illustrated in \reffig{fig:rrrspeed}, all controllers, with the exception of the \ac{PD} controller, demonstrated the ability to maintain the maximum allowable relative velocity with a high degree of precision. Nevertheless, both the \ac{PD} and \ac{CTM} controllers are restricted by the effective mass of the robot, as estimated by the online system. The $\text{IMP}_1$ controller, owing to the reduction in effective mass of the manipulator, exhibited a slightly higher velocity threshold compared to the preceding two controllers and was able to maintain this velocity reliably until the settling time, beyond which it gradually decreased its speed until coming to a halt. The second impedance controller $\text{IMP}_2$ achieves and sustains higher speeds during the trajectory tracking. Subsequently, this results in lead-time reduction, which is beneficial in industrial applications.

\begin{figure}[t!]
\centering
\subfloat{\includegraphics[width=0.9\linewidth]{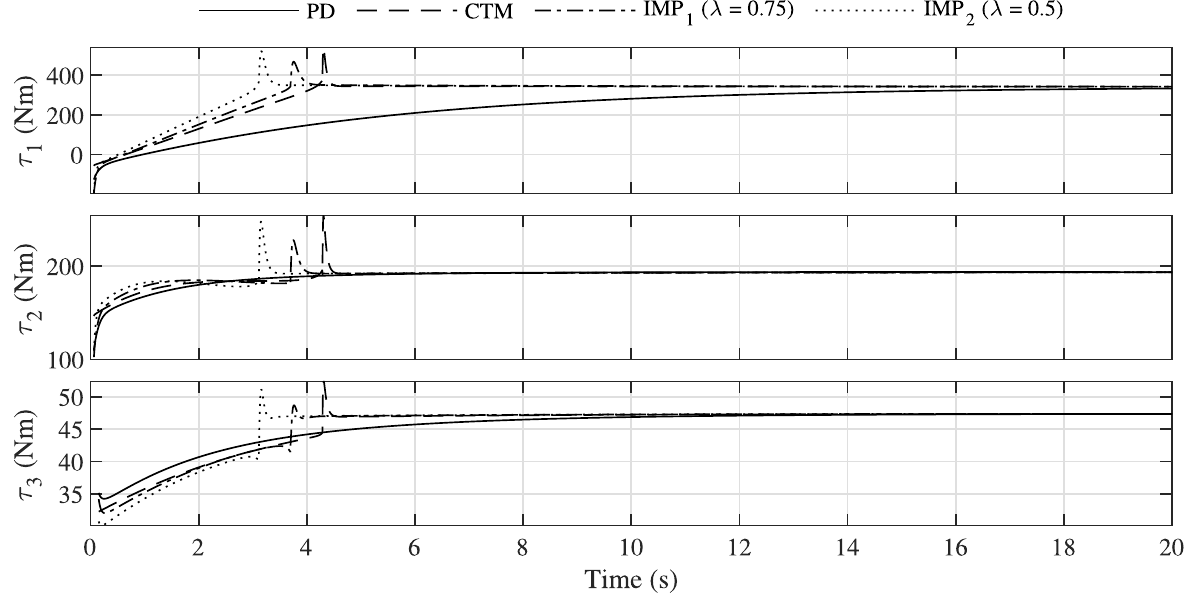}}
\caption{Instantaneous joint input torques of 3R robot.}
\label{fig:rrrtau}
\end{figure}

In \reffig{fig:rrrtau}, the torque values applied to each actuator of the robot are presented to illustrate the level of control effort. The cumulative absolute torque value is used for comparative analysis of controller performance, taking into account the degree of error minimization and the level of control effort required to adhere to the desired trajectory. The evaluated controllers---including \ac{PD}, \ac{CTM}, $\text{IMP}_1$, and $\text{IMP}_2$---show control efforts of 9495, 10,793, 10,921, and 11,049 Nm$\cdot$s, respectively. The $\text{IMP}_2$ controller, which demonstrated the most effective reduction in error and lead time, required only a modest $16\%$ increase in total control effort compared to the minimum value reported for the \ac{PD} controller. This slight increase in control effort is considered a reasonable trade-off given the significant benefits previously discussed.

\subsection{Franka Emika Panda Robot with 3D Workspace Control}

\begin{figure}[t!]
\centering
\subfloat[\centering]{\includegraphics[height=0.4\linewidth]{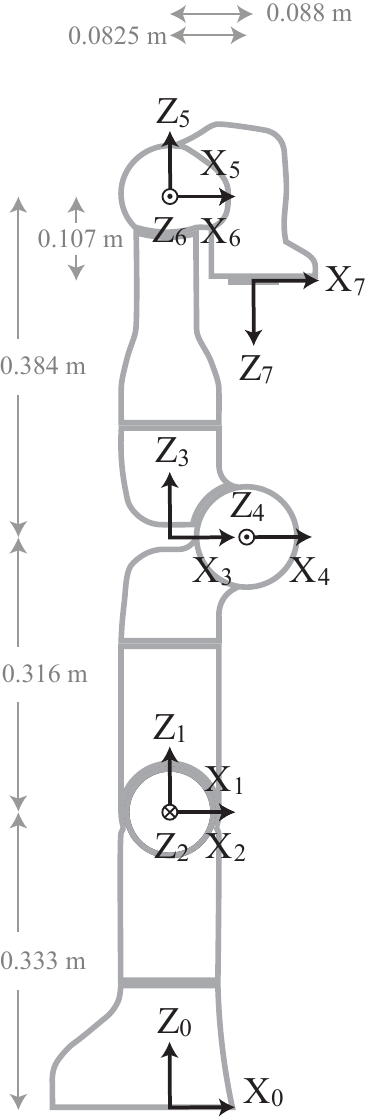}}
\quad
\subfloat[\centering]{\includegraphics[height=0.4\linewidth]{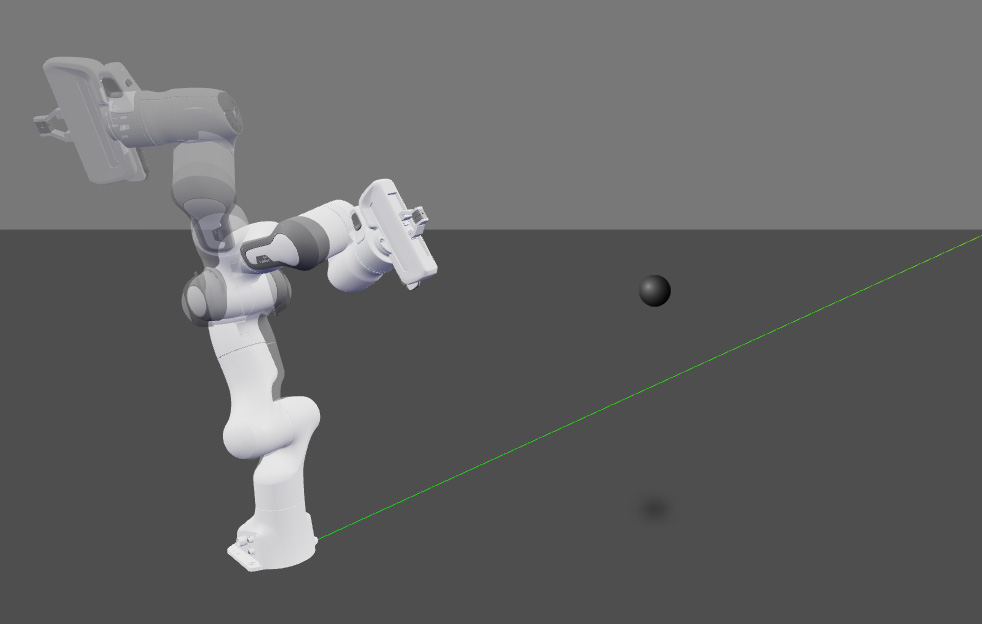}}
\caption{(a) Franka-Emika Panda serial-link manipulator in its zero angle configurations. (b) Visualisation of a Panda robot that has been controlled by
the proposed variable impedance control scheme (semi-transparent robot: initial configuration; opaque robot: final configuration). The black sphere depicts the position of a generic obstacle in the workspace that is stationary during robot movement. The obstacle is taken to be the face region of a human operator for simulation purposes.}
\label{fig:panda}
\end{figure}

The Franka Emika Panda robot is a cutting-edge robotic system designed for advanced industrial and research applications and is depicted in \reffig{fig:panda}. It is a highly flexible, seven-axis robot with a lightweight, compact design that allows it to perform intricate tasks with precision and accuracy. Equipped with advanced sensors and control algorithms, the Panda robot can adapt to different environments and is used to interact with humans. It offers a wide range of capabilities, including pick-and-place operations, assembly tasks, and delicate manipulation of objects.

The manipulator under consideration exhibits a maximum reach of 800 mm and has a payload capacity of up to 7 kg, thus rendering it applicable to a wide range of contexts. It functions within the 3D coordinate space and is characterized by seven \ac{DoF}. Specifically, all seven actuators in the manipulator are revolution joints, with the vector of angular values denoted by $Q = [q_1, q_2, q_3, q_4, q_5, q_6, q_7]^T \in {\mathbb{R}^7}$. To describe the position and orientation of the end-effector in the task space, six coordinates are required. In the present simulation, the orientation of the end-effector is represented by the roll--pitch--yaw angles, and the rotations are effected through the ZYX convention. These rotations are anchored on fixed global axes located at the robot base (i.e., extrinsic rotation axis). Accordingly, the generalized coordinate vector of the task space pose can be expressed as $r = [x, y, z, \alpha, \beta, \gamma]^T \in {\mathbb{R}^6}$. The Denavit--Hartenberg frames for this robot are shown in \reffig{fig:panda}, alongside the table of parameters in \reftab{tab:pandadh}. The dynamic parameters of the robot (inertia matrix, gravity vector, Coriolis term, and Jacobian) are calculated numerically using the package provided in \cite{gaz2019dynamic}. 

\begin{table}[!t]
\caption{Denavit-Hartenberg parameters for the Franka Emika Panda \cite{gaz2019dynamic}. The reference frames follow the modified Denavit-Hartenberg convention. In the table, $d_1$ = 0.333 m, $d_3$ = 0.316 m, $d_5$ = 0.384 m, $d_f$ = 0.107 m, $a_4$ = 0.0825 m, $a_5$ = -0.0825 m, $a_7$ = 0.088 m}
\centering
\begin{tabularx}{\textwidth}{@{} >{\centering\arraybackslash}X >{\centering\arraybackslash}X >{\centering\arraybackslash}X >{\centering\arraybackslash}X >{\centering\arraybackslash}X @{}}
\toprule
$\pmb{i}$ & $\pmb{a_i}$ & $\pmb{\alpha_i}$ & $\pmb{d_i}$ & $\pmb{\theta_i}$ \\
\midrule
1 & 0 & 0 & $d_1$ & $q_1$ \\
2 & 0 & $- \pi / 2$ & 0 & $q_2$ \\
3 & 0 & $ \pi / 2$ & $d_3$ & $q_3$ \\
4 & $a_4$ & $ \pi / 2$ & 0 & $q_4$ \\
5 & $a_5$ & $- \pi / 2$ & $d_5$ & $q_5$ \\
6 & 0 & $ \pi / 2$ & 0 & $q_6$ \\
7 & $a_7$ & $ \pi / 2$ & 0 & $q_7$ \\
8 & 0 &0 & $d_f$ & 0 \\
\bottomrule
\end{tabularx}
\label{tab:pandadh}
\end{table}

To compare the efficacy of the proposed control scheme with conventional controllers, the task of moving the end-effector coordinates from the initial pose $r_{start} = [0.18, -0.36, 0.98, 0.49, -0.77, 1.72]$ to the target pose $r_{final} = [0.33, 0.51, 0.52, 2.8, 0.52$, $-1.69]$ has been assigned. Similar to the 3R robot considered in the preceding section, each controller must ensure that the robot is operated at the highest safe velocity in compliance with the prevailing standards. It is assumed that a human operator is stationary at position $X_\text{Human} = [0.45, 0.65, 0.70]$, and there is a risk of the robot's end-effector colliding with the operator's ``Face'' region. The pertinent parameters required for calculating the maximum safe velocity in accordance with the standards are provided in \reftab{tab:1}.

\begin{figure}[t!]
\centering
\subfloat{\includegraphics[width=0.9\linewidth]{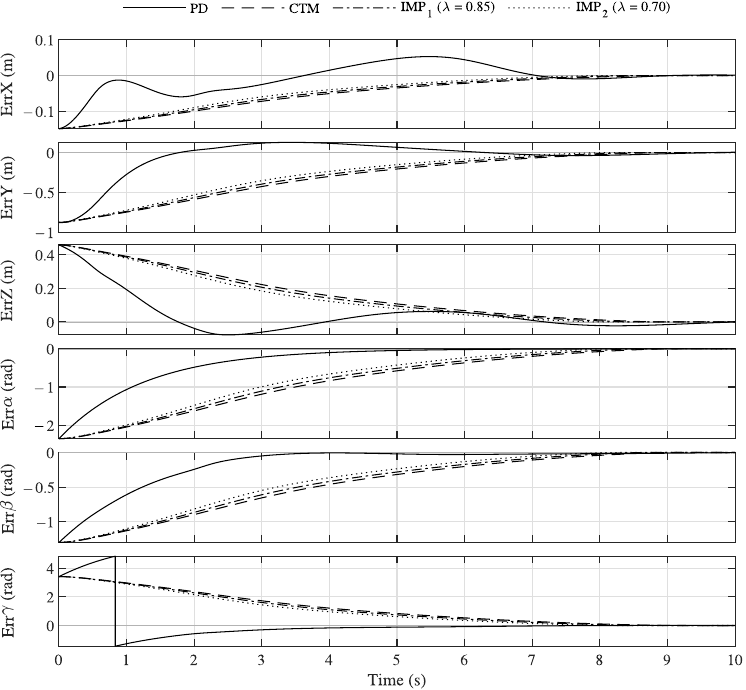}}
\caption{Instantaneous tracking errors of workspace coordinates of Franka Emika Panda robot.}
\label{fig:pandaerr}
\end{figure}

\begin{figure}[t!]
\centering
\subfloat{\includegraphics[width=0.9\linewidth]{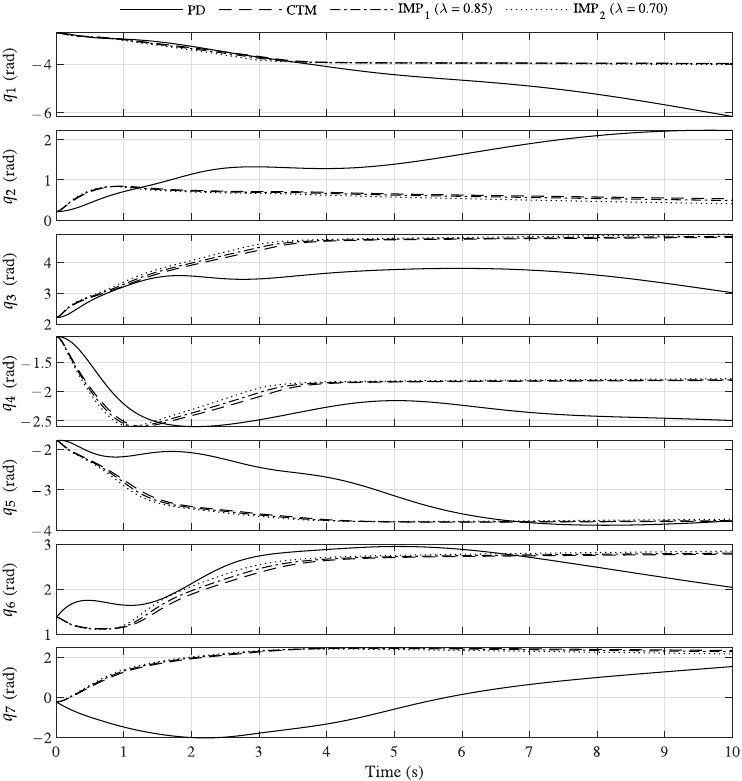}}
\caption{Instantaneous trajectory of the Franka Emika Panda robot end-effector in joint space coordinates.}
\label{fig:pandatraj}
\end{figure}

Similar to the previous simulation, in order to keep all experiments consistent and to be able to compare the results in a comprehensive and meaningful manner, the control parameters 
\begin{equation}
  {K_p} = \left( {\begin{array}{*{5}{c}}
    {{{20}_{3 \times 3}}}&{{0_{3 \times 3}}} \\ 
    {{0_{3 \times 3}}}&{{{5}_{3 \times 3}}} 
  \end{array}} \right), 
  \end{equation}
  and
  \begin{equation}
  {K_d} = \left( {\begin{array}{*{20}{c}}
    {2{{\sqrt {20} }_{3 \times 3}}}&{{0_{3 \times 3}}} \\ 
    {{0_{3 \times 3}}}&{2{{\sqrt {5} }_{3 \times 3}}}
  \end{array}} \right), 
  \end{equation}
used for each controller are set to the same values. The selection of these values was undertaken through an empirical process involving the assessment of system performance across a range of $K_p$ values spanning from 5 to 50 in increments of 5. The value for $K_d$ is set as $2\sqrt{K_p}$ so that the resulting system has a damping ratio of approximately 1, resulting in a critically damped performance, which is desirable in many control applications. The performance of the impedance controller is assessed via two distinct simulations, in which $\lambda$ takes on the values of 0.80 ($\text{IMP}_1$) and 0.60 ($\text{IMP}_2$), in conjunction with the \ac{CTM} and \ac{PD} controllers. Each of these controllers is applied to the Panda robot, and the ensuing tracking error in Cartesian coordinates of the end-effector is depicted in \reffig{fig:pandaerr}. Also, the instantaneous joint-space trajectory is show in \reffig{fig:pandatraj}. 

Settling time and tracking errors are computed and presented in \reftab{tab:pandareport}. The \ac{PD}, \ac{CTM}, $\text{IMP}_1$, and $\text{IMP}_2$ controllers settled in $8.1775$, $7.4779$, $7.1471$, and $6.7022$ seconds, respectively. From the findings, it can be observed that $\text{IMP}_1$ exhibited a $12.6\%$ improvement in settling time over the \ac{PD} controller and a $4.4\%$ improvement over the \ac{CTM} controller. Similarly, $\text{IMP}_2$ showed greater improvement, with a reduction in settling time of $18\%$ compared to the \ac{PD} controller and $10.3\%$ compared to the \ac{CTM} controller.

\begin{figure}[t!]
\centering
\subfloat[\centering PD Controller]{\includegraphics[width=0.45\linewidth]{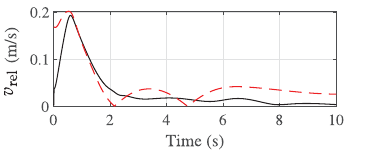}} \quad
\subfloat[\centering CTM Controller]{\includegraphics[width=0.45\linewidth]{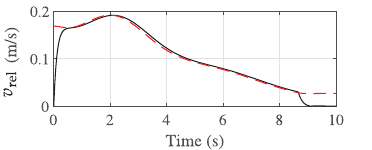}} 
\
\subfloat[\centering $\text{IMP}_1$ Controller]{\includegraphics[width=0.45\linewidth]{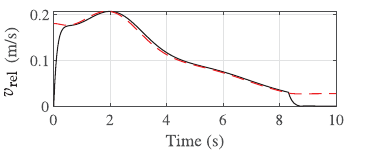}} \quad
\subfloat[\centering $\text{IMP}_2$ Controller]{\includegraphics[width=0.45\linewidth]{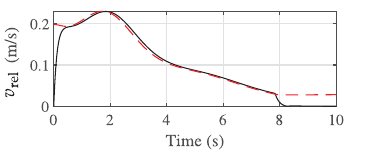}}
\caption{Instantaneous controller relative velocity of Franka Emika Panda robot. (dashed red: ISO/TS 15066 PFL maximum velocity; solid: actual velocity)}
\label{fig:pandavel}
\end{figure}

The results demonstrate that a reduction in the $\lambda$ value leads to a decrease in the mean tracking error for all task-space coordinates, except for the \ac{PD} controller, which exhibits a lower mean error. However, the \ac{PD} controller suffers from oscillatory behavior, resulting in a higher settling time. Furthermore, it lacks the capability to specify a desired velocity during control, making it challenging to reliably control the robot under the velocity threshold set by the standard. This issue is particularly pronounced when evaluating the robot's relative velocity concerning the maximum allowable velocity, as depicted in \reffig{fig:pandavel}.

\begin{table}[t!]
\caption{Comparison between the performance of simulated controllers of Franka Emika Panda robot.}
\centering
\begin{tabularx}{\textwidth}{@{} X >{\centering\arraybackslash}X>{\centering\arraybackslash}X>{\centering\arraybackslash}X>{\centering\arraybackslash}X >{\centering\arraybackslash}X>{\centering\arraybackslash}X>{\centering\arraybackslash}X@{}}
\toprule
\textbf{} & \textbf{Settling time} (s) & \multicolumn{6}{c}{\textbf{Average tracking error in task-space}} \\ 
\cmidrule(lr){3-8}
 & & $\mathrm{\mathbf{{\bar e}_x}}$ (m) & $\mathrm{\mathbf{{\bar e}_y}}$ (m) & $\mathrm{\mathbf{{\bar e}_z}}$ (m) & $\mathrm{\mathbf{{\bar e}}}_{\boldsymbol{\alpha}}$ (rad) & $\mathrm{\mathbf{{\bar e}}}_{\boldsymbol{\beta}}$ (rad)& $\mathrm{\mathbf{{\bar e}}}_{\boldsymbol{\gamma}}$ (rad)\\
\midrule
\textbf{PD} & 8.1775 & 0.0280 & 0.1136 & 0.0670 & 0.2972 & 0.1566 & 0.5580 \\
\textbf{CTM} & 7.4779 & 0.0493 & 0.2911 & 0.1529 & 0.8080 & 0.4453 & 1.1741 \\
\textbf{$\text{IMP}_1$} & 7.1471 & 0.0466 & 0.2747 & 0.1442 & 0.7641 & 0.4210 & 1.1102 \\
\textbf{$\text{IMP}_2$} & 6.7022 & 0.0429 & 0.2532 & 0.1330 & 0.7065 & 0.3893 & 1.0266  \\ 
\bottomrule
\end{tabularx}
\label{tab:pandareport}
\end{table}

\begin{figure}[t!]
\centering
\subfloat{\includegraphics[width=0.9\linewidth]{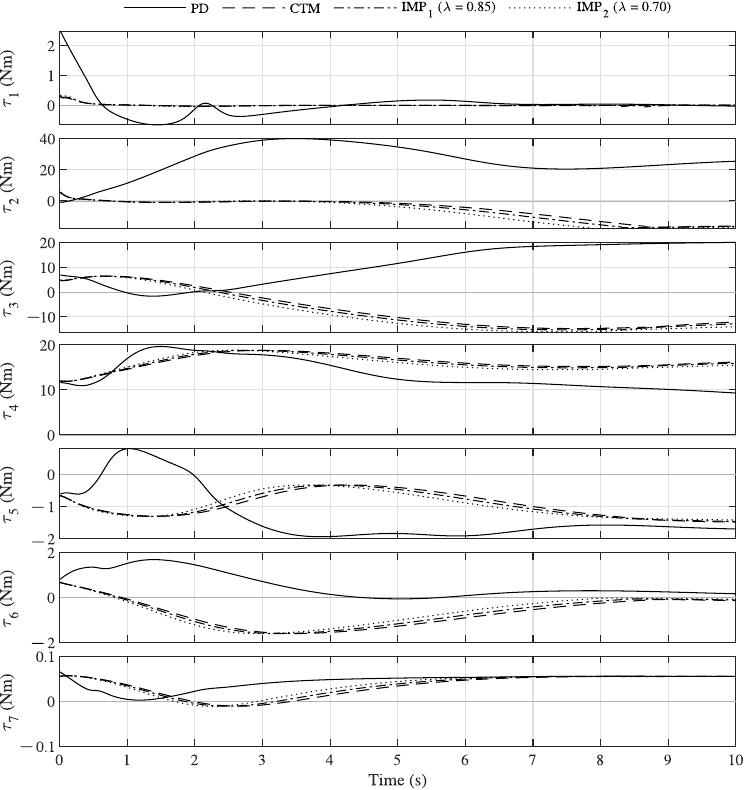}}
\caption{Instantaneous joint input torques of Franka Emika Panda robot.}
\label{fig:pandatau}
\end{figure}

\reffig{fig:pandatau} illustrates the torque values of the seven actuators of the panda robot. During simulation, the torque limits for each actuator are enforced based on the manufacturer's specifications. Specifically, the first four actuators have a torque limit of 87 Nm, while the last three have a limit of 12 Nm. It is also noteworthy to observe that the torque rate limit assigned to each of the joint actuators by the manufacturer is $\dot{\tau}_{max} = 1000$~Nm/s. This assertion is visually validated by examining \reffig{fig:pandatau}, where it becomes evident that the simulated control mechanisms operate significantly below this specified threshold.

\section{Conclusions}
\label{sec:6}

This paper has investigated variable impedance control design for robotic systems to enable safe operation at higher speeds while complying with safety standards. Three methods for determining effective mass have been presented and compared in various simulation scenarios. According to achieved results, the proposed impedance control scheme significantly increases the maximum permissible velocity of two robotic manipulators compared to alternative control strategies while limiting contact forces to below safety thresholds. The proposed controller has achieved this by allowing the effective mass of the robotic system to be adjusted through impedance control. The reduction factor $\lambda$ has been introduced to effectively reduce the mass, which improves the robot's responsiveness. The proposed controller offers a practical and efficient approach to robotic manipulation in scenarios that require high-speed operation while also ensuring safety. However, certain aspects need to be explored in future work. Investigation of the closed-loop stability of the proposed control scheme must be addressed.  Moreover, advanced optimization methodologies can be employed to finely calibrate the parameter  $\lambda$ in order to align it precisely with the requirements of the given task. Also, it is imperative to evaluate the efficacy of the controller through experimentation on a test platform. This endeavor would facilitate the examination of the control framework within real-world operational scenarios, incorporating interaction and impact dynamics between the robotic system and human~collaborator.

\bibliographystyle{unsrt}  
\bibliography{sn-bibliography} 

\begin{thebibliography}{10}

\bibitem{ISOTS15066}
{ISO/TS} 15066:2016. {R}obots and robotic devices - {C}ollaborative robots.
\newblock Technical specification, International Organization for
  Standardization, Geneva, Switzerland, 2016.

\bibitem{ISO102181}
{ISO} 10218-1:2011. {R}obots and robotic devices - {S}afety requirements for
  industrial robots: {R}obots.
\newblock Standard, International Organization for Standardization, Geneva,
  Switzerland, 2011.

\bibitem{ISO102182}
{ISO} 10218-2:2011. {R}obots and robotic devices - {S}afety requirements for
  industrial robots: {R}obot systems and integration.
\newblock Standard, International Organization for Standardization, Geneva,
  Switzerland, 2011.

\bibitem{ISO13482}
{ISO} 13482:2014. {R}obots and robotic devices — {S}afety requirements for
  personal care robots.
\newblock Standard, International Organization for Standardization, Geneva,
  Switzerland, 2014.

\bibitem{ogorodnikova2008methodology}
Olesya Ogorodnikova.
\newblock Methodology of safety for a human robot interaction designing stage.
\newblock In {\em IEEE Conference on Human System Interactions}, pages
  452--457, 2008.

\bibitem{long2017industrial}
Philip Long, Christine Chevallereau, Damien Chablat, and Alexis Girin.
\newblock An industrial security system for human-robot coexistence.
\newblock {\em Industrial Robot: An International Journal}, 2017.

\bibitem{karagiannis2022adaptive}
Panagiotis Karagiannis, Niki Kousi, George Michalos, Konstantinos Dimoulas,
  Konstantinos Mparis, Dimosthenis Dimosthenopoulos, {\"O}nder Tok{\c{c}}alar,
  Toni Guasch, Gian~Paolo Gerio, and Sotiris Makris.
\newblock Adaptive speed and separation monitoring based on switching of safety
  zones for effective human robot collaboration.
\newblock {\em Robotics and Computer-Integrated Manufacturing}, 77:102361,
  2022.

\bibitem{scalera2020application}
Lorenzo Scalera, Andrea Giusti, Renato Vidoni, V~Di~Cosmo, Dominik Matt, and
  Michael Riedl.
\newblock Application of dynamically scaled safety zones based on the {ISO/TS}
  15066:2016 for collaborative robotics.
\newblock {\em International Journal of Mechanics and Control}, 21(1):41--49,
  2020.

\bibitem{kim2020estimating}
Eugene Kim, Robin Kirschner, Yoji Yamada, and Shogo Okamoto.
\newblock Estimating probability of human hand intrusion for speed and
  separation monitoring using interference theory.
\newblock {\em Robotics and Computer-Integrated Manufacturing}, 61:101819,
  2020.

\bibitem{safeea2019minimum}
Mohammad Safeea and Pedro Neto.
\newblock Minimum distance calculation using laser scanner and {IMU}s for safe
  human-robot interaction.
\newblock {\em Robotics and Computer-Integrated Manufacturing}, 58:33--42,
  2019.

\bibitem{halme2018review}
Roni-Jussi Halme, Minna Lanz, Joni K{\"a}m{\"a}r{\"a}inen, Roel Pieters, Jyrki
  Latokartano, and Antti Hietanen.
\newblock Review of vision-based safety systems for human-robot collaboration.
\newblock {\em Procedia Cirp}, 72:111--116, 2018.

\bibitem{choi2022integrated}
Sung~Ho Choi, Kyeong-Beom Park, Dong~Hyeon Roh, Jae~Yeol Lee, Mustafa Mohammed,
  Yalda Ghasemi, and Heejin Jeong.
\newblock An integrated mixed reality system for safety-aware human-robot
  collaboration using deep learning and digital twin generation.
\newblock {\em Robotics and Computer-Integrated Manufacturing}, 73:102258,
  2022.

\bibitem{seemann2004head}
Edgar Seemann, Kai Nickel, and Rainer Stiefelhagen.
\newblock Head pose estimation using stereo vision for human-robot interaction.
\newblock In {\em sixth IEEE International Conference on Automatic Face and
  Gesture Recognition}, pages 626--631, 2004.

\bibitem{li2023ar}
Chengxi Li, Pai Zheng, Yue Yin, Yat~Ming Pang, and Shengzeng Huo.
\newblock An {AR}-assisted deep reinforcement learning-based approach towards
  mutual-cognitive safe human-robot interaction.
\newblock {\em Robotics and Computer-Integrated Manufacturing}, 80:102471,
  2023.

\bibitem{hietanen2020ar}
Antti Hietanen, Roel Pieters, Minna Lanz, Jyrki Latokartano, and Joni-Kristian
  K{\"a}m{\"a}r{\"a}inen.
\newblock {AR}-based interaction for human-robot collaborative manufacturing.
\newblock {\em Robotics and Computer-Integrated Manufacturing}, 63:101891,
  2020.

\bibitem{khatib2021human}
Maram Khatib, Khaled Al~Khudir, and Alessandro De~Luca.
\newblock Human-robot contactless collaboration with mixed reality interface.
\newblock {\em Robotics and Computer-Integrated Manufacturing}, 67:102030,
  2021.

\bibitem{oliff2020reinforcement}
Harley Oliff, Ying Liu, Maneesh Kumar, Michael Williams, and Michael Ryan.
\newblock Reinforcement learning for facilitating human-robot-interaction in
  manufacturing.
\newblock {\em Journal of Manufacturing Systems}, 56:326--340, 2020.

\bibitem{gao2020robust}
Qing Gao, Jinguo Liu, and Zhaojie Ju.
\newblock Robust real-time hand detection and localization for space
  human--robot interaction based on deep learning.
\newblock {\em Neurocomputing}, 390:198--206, 2020.

\bibitem{el2020towards}
Mohamed El-Shamouty, Xinyang Wu, Shanqi Yang, Marcel Albus, and Marco~F Huber.
\newblock Towards safe human-robot collaboration using deep reinforcement
  learning.
\newblock In {\em IEEE International Conference on Robotics and Automation},
  pages 4899--4905, 2020.

\bibitem{magrini2015control}
Emanuele Magrini, Fabrizio Flacco, and Alessandro De~Luca.
\newblock Control of generalized contact motion and force in physical
  human-robot interaction.
\newblock In {\em IEEE international conference on robotics and automation},
  pages 2298--2304, 2015.

\bibitem{papanastasiou2019towards}
Stergios Papanastasiou, Niki Kousi, Panagiotis Karagiannis, Christos
  Gkournelos, Apostolis Papavasileiou, Konstantinos Dimoulas, Konstantinos
  Baris, Spyridon Koukas, George Michalos, and Sotiris Makris.
\newblock Towards seamless human robot collaboration: integrating multimodal
  interaction.
\newblock {\em The International Journal of Advanced Manufacturing Technology},
  105:3881--3897, 2019.

\bibitem{villani2020wearable}
Valeria Villani, Massimiliano Righi, Lorenzo Sabattini, and Cristian Secchi.
\newblock Wearable devices for the assessment of cognitive effort for
  human--robot interaction.
\newblock {\em IEEE Sensors Journal}, 20(21):13047--13056, 2020.

\bibitem{dimitropoulos2021seamless}
Nikos Dimitropoulos, Theodoros Togias, Natalia Zacharaki, George Michalos, and
  Sotiris Makris.
\newblock Seamless human--robot collaborative assembly using artificial
  intelligence and wearable devices.
\newblock {\em Applied Sciences}, 11(12):5699, 2021.

\bibitem{mohammadi2020mixed}
Fatemeh Mohammadi~Amin, Maryam Rezayati, Hans~Wernher van~de Venn, and Hossein
  Karimpour.
\newblock A mixed-perception approach for safe human--robot collaboration in
  industrial automation.
\newblock {\em Sensors}, 20(21):6347, 2020.

\bibitem{byner2019dynamic}
Christoph Byner, Bj{\"o}rn Matthias, and Hao Ding.
\newblock Dynamic speed and separation monitoring for collaborative robot
  applications--concepts and performance.
\newblock {\em Robotics and Computer-Integrated Manufacturing}, 58:239--252,
  2019.

\bibitem{rosenstrauch2018human}
Martin~J Rosenstrauch, Tessa~J Pannen, and J{\"o}rg Kr{\"u}ger.
\newblock Human robot collaboration-using {Kinect} {V2} for {ISO/TS} 15066
  speed and separation monitoring.
\newblock {\em Procedia CIRP}, 76:183--186, 2018.

\bibitem{kang2022manipulator}
Yeon Kang, Donghan Kim, and Dongwon Yun.
\newblock Manipulator collision avoidance system based on a 3{D} potential
  field with {ISO} 15066.
\newblock {\em IEEE Access}, 2022.

\bibitem{du2018active}
Guanglong Du, Shuaiying Long, Fang Li, and Xin Huang.
\newblock Active collision avoidance for human-robot interaction with {UKF},
  expert system, and artificial potential field method.
\newblock {\em Frontiers in Robotics and AI}, 5:125, 2018.

\bibitem{vysocky2019motion}
Ale{\v{s}} Vysock{\`y}, Hisaka Wada, Jun Kinugawa, and Kazuhiro Kosuge.
\newblock Motion planning analysis according to {ISO/TS} 15066 in human--robot
  collaboration environment.
\newblock In {\em IEEE/ASME International Conference on Advanced Intelligent
  Mechatronics}, pages 151--156, 2019.

\bibitem{shin2018allowable}
Heonseop Shin, Kwang Seo, and Sungsoo Rhim.
\newblock Allowable maximum safe velocity control based on human-robot distance
  for collaborative robot.
\newblock In {\em 15th International Conference on Ubiquitous Robots}, pages
  401--405, 2018.

\bibitem{aivaliotis2019power}
Panagiotis Aivaliotis, S~Aivaliotis, Christos Gkournelos, K~Kokkalis, George
  Michalos, and Sotiris Makris.
\newblock Power and force limiting on industrial robots for human-robot
  collaboration.
\newblock {\em Robotics and Computer-Integrated Manufacturing}, 59:346--360,
  2019.

\bibitem{ferraguti2020control}
Federica Ferraguti, Mattia Bertuletti, Chiara~Talignani Landi, Marcello
  Bonf{\`e}, Cesare Fantuzzi, and Cristian Secchi.
\newblock A control barrier function approach for maximizing performance while
  fulfilling to {ISO/TS} 15066 regulations.
\newblock {\em IEEE Robotics and Automation Letters}, 5(4):5921--5928, 2020.

\bibitem{lucci2020combining}
Niccol{\`o} Lucci, Bakir Lacevic, Andrea~Maria Zanchettin, and Paolo Rocco.
\newblock Combining speed and separation monitoring with power and force
  limiting for safe collaborative robotics applications.
\newblock {\em IEEE Robotics and Automation Letters}, 5(4):6121--6128, 2020.

\bibitem{yoon2005safe}
Seong-Sik Yoon, Sungchul Kang, Seung-kook Yun, Seung-Jong Kim, Young-Hwan Kim,
  and Munsang Kim.
\newblock Safe arm design with mr-based passive compliant joints and
  visco—elastic covering for service robot applications.
\newblock {\em Journal of mechanical science and technology}, 19:1835--1845,
  2005.

\bibitem{park2008safe}
Jung-Jun Park, Byeong-Sang Kim, Jae-Bok Song, and Hong-Seok Kim.
\newblock Safe link mechanism based on nonlinear stiffness for collision
  safety.
\newblock {\em Mechanism and Machine Theory}, 43(10):1332--1348, 2008.

\bibitem{wolf2008new}
Sebastian Wolf and Gerd Hirzinger.
\newblock A new variable stiffness design: Matching requirements of the next
  robot generation.
\newblock In {\em 2008 IEEE International Conference on Robotics and
  Automation}, pages 1741--1746. IEEE, 2008.

\bibitem{seriani2018development}
Stefano Seriani, Paolo Gallina, Lorenzo Scalera, and Vanni Lughi.
\newblock Development of n-dof preloaded structures for impact mitigation in
  cobots.
\newblock {\em Journal of Mechanisms and Robotics}, 10(5):051009, 2018.

\bibitem{martinetti2021redefining}
Alberto Martinetti, Peter~K Chemweno, Kostas Nizamis, and Eduard
  Fosch-Villaronga.
\newblock Redefining safety in light of human-robot interaction: A critical
  review of current standards and regulations.
\newblock {\em Frontiers in chemical engineering}, 3:32, 2021.

\bibitem{khatib1995inertial}
Oussama Khatib.
\newblock Inertial properties in robotic manipulation: An object-level
  framework.
\newblock {\em The international journal of robotics research}, 14(1):19--36,
  1995.

\bibitem{hogan1984impedance}
Neville Hogan.
\newblock Impedance control: An approach to manipulation.
\newblock In {\em American Control Conference}, pages 304--313, 1984.

\bibitem{hogan1985impedance}
Neville Hogan.
\newblock {Impedance Control: An Approach to Manipulation: Part
  II—Implementation}.
\newblock {\em Journal of Dynamic Systems, Measurement, and Control},
  107(1):8--16, 03 1985.

\bibitem{dragan2015effects}
Anca~D Dragan, Shira Bauman, Jodi Forlizzi, and Siddhartha~S Srinivasa.
\newblock Effects of robot motion on human-robot collaboration.
\newblock In {\em Tenth Annual ACM/IEEE International Conference on Human-Robot
  Interaction}, pages 51--58, 2015.

\bibitem{hoffman2019evaluating}
Guy Hoffman.
\newblock Evaluating fluency in human--robot collaboration.
\newblock {\em IEEE Transactions on Human-Machine Systems}, 49(3):209--218,
  2019.

\bibitem{scalera2022enhancing}
Lorenzo Scalera, Andrea Giusti, Renato Vidoni, and Alessandro Gasparetto.
\newblock Enhancing fluency and productivity in human-robot collaboration
  through online scaling of dynamic safety zones.
\newblock {\em The International Journal of Advanced Manufacturing Technology},
  121(9-10):6783--6798, 2022.

\bibitem{rtb}
Peter Corke and Jesse Haviland.
\newblock Not your grandmother’s toolbox--the robotics toolbox reinvented for
  python.
\newblock In {\em IEEE International Conference on Robotics and Automation},
  pages 11357--11363, 2021.

\bibitem{gaz2019dynamic}
Claudio Gaz, Marco Cognetti, Alexander Oliva, Paolo~Robuffo Giordano, and
  Alessandro De~Luca.
\newblock Dynamic identification of the franka emika panda robot with retrieval
  of feasible parameters using penalty-based optimization.
\newblock {\em IEEE Robotics and Automation Letters}, 4(4):4147--4154, 2019.

\end{thebibliography}

\end{document}